\documentclass[a4paper,fleqn]{cas-dc}

\usepackage[numbers]{natbib}
\hypersetup{ colorlinks, citecolor=blue, linkcolor=blue, urlcolor=Blue}
\def\tsc#1{\csdef{#1}{\textsc{\lowercase{#1}}\xspace}}
\tsc{WGM}
\tsc{QE}
\tsc{EP}
\tsc{PMS}
\tsc{BEC}
\tsc{DE}

\usepackage{lineno}
\usepackage{multirow}
\usepackage{color,soul}

\newcommand{\highlighting}[1]{%
    {%
    \sethlcolor{white}%
    \hl{#1}%
    }%
}

\begin{document}
\let\WriteBookmarks\relax
\def\floatpagepagefraction{1}
\def\textpagefraction{.001}

\shorttitle{PolSAR Image Classification Using DDF2Pol}

\shortauthors{M.Q. Alkhatib}



\title [mode = title]{DDF2Pol: A Dual-Domain Feature Fusion Network for PolSAR Image Classification}

\author[1]{Mohammed Q. Alkhatib}[orcid=0000-0003-4812-614X
]

\ead{mqalkhatib@ieee.org}

\ead[url]{https://www.linkedin.com/in/mqalkhatib85/}

\credit{Conceptualization of this study, Methodology, Software, Data curation, Writing - Original draft preparation}

\affiliation[1]{organization={College of Engineering and IT, University of Dubai},
            city={Dubai},
            postcode={14143}, 
            country={UAE}}

\begin{abstract}
This paper presents DDF2Pol, a lightweight dual-domain convolutional neural network for PolSAR image classification. The proposed architecture integrates two parallel feature extraction streams—one real-valued and one complex-valued—designed to capture complementary spatial and polarimetric information from PolSAR data. To further refine the extracted features, a depth-wise convolution layer is employed for spatial enhancement, followed by a coordinate attention mechanism to focus on the most informative regions. Experimental evaluations conducted on two benchmark datasets, Flevoland and San Francisco, demonstrate that DDF2Pol achieves superior classification performance while maintaining low model complexity. Specifically, it attains an Overall Accuracy (OA) of 98.16\% on the Flevoland dataset and 96.12\% on the San Francisco dataset, outperforming several state-of-the-art real- and complex-valued models. With only 91,371 parameters, DDF2Pol offers a practical and efficient solution for accurate PolSAR image analysis, even when training data is limited. The source code is publicly available at \url{https://github.com/mqalkhatib/DDF2Pol}

\end{abstract}




\maketitle

\section{Introduction}
\label{intro}
Polarimetric Synthetic Aperture Radar (PolSAR) is a powerful remote sensing technology capable of acquiring detailed information about surface structures under all-weather and day-and-night conditions. By transmitting and receiving electromagnetic waves across multiple polarization states, PolSAR systems offer a richer representation of targets than conventional SAR systems. These capabilities make PolSAR widely applicable in fields such as urban mapping \cite{zhang2016mapping}, geospatial monitoring \cite{ponnurangam2018application}, agriculture, and environmental assessment \citep{tebege2025geospatial}. A fundamental task in PolSAR analysis is classification, which involves assigning each pixel to a land cover category. Accurate classification is critical for understanding the structure and dynamics of observed scenes.

Traditionally, PolSAR classification relied on machine learning methods such as Support Vector Machines (SVMs) \citep{lardeux2009support} and decision trees \citep{qi2012novel}. These methods depend on hand-crafted features derived from polarimetric decomposition or statistical measures, which often fall short in capturing the full complexity of PolSAR data. In recent years, deep learning—particularly convolutional neural networks (CNNs)—has become popular due to its ability to automatically learn useful features directly from raw data \citep{zhou2016polarimetric}. CNNs generally outperform classical approaches, especially when sufficient labeled data is available. To better handle the spatial and polarimetric structure of PolSAR images, researchers have introduced 3D convolutional layers \citep{zhang2018polarimetric} and hybrid architectures. \highlighting{Unlike traditional 2D CNNs, which operate only across spatial dimensions, 3D convolutions process both spatial and channel (polarimetric) dimensions at once. This allows the model to learn richer features from local regions while keeping the number of parameters lower. As a result, 3D CNNs help reduce overfitting and often lead to better classification performance, particularly when training data is limited.}

While early CNN\highlighting{s}-based models focused on real-valued data, they often ignored the phase information present in complex-valued PolSAR images. However, studies have shown that phase components carry valuable scattering information that can enhance classification accuracy \citep{zhang2017complex}. This has led to the development of complex-valued CNNs (CV-CNNs), which process both magnitude and phase directly and have demonstrated improved performance over their real-valued counterparts \citep{tan2019complex}. More advanced models have combined the strengths of real and complex-valued networks, aiming to extract richer representations while balancing model complexity and computational cost.

To further improve feature discrimination, attention mechanisms have been introduced in both CNN\highlighting{s} and CV-CNN\highlighting{s} frameworks. These mechanisms allow models to focus on the most relevant spatial and polarimetric features while suppressing less informative ones \citep{hu2018squeeze, hou2021coordinate}. Attention modules such as squeeze-and-excitation blocks and coordinate attention have proven effective in enhancing classification results across multiple PolSAR datasets. Recent works also explore attention in multiscale or sequential architectures, demonstrating the potential of adaptive weighting strategies for more robust learning \citep{kong2023multi, yang2021composite}.

Real-valued CNNs, when applied to engineered descriptors, offer computational efficiency and have shown promising performance. On the other hand, complex-valued networks extract amplitude and phase information directly from raw PolSAR data, capturing richer scattering features. To benefit from both domains, this paper proposes a dual-domain model that processes and fuses features from real and complex representations. By combining their strengths and refining the fused output using attention and depth-wise convolution, the proposed model enhances classification performance while remaining lightweight and efficient.

\highlighting{Recent advances in PolSAR classification have explored feature fusion, efficient architectures, and decision-level strategies. Decomposition-based fusion techniques have combined coherent and non-coherent polarimetric features to improve classification through complementary scattering information} \cite{karachristos2024fusion}. \highlighting{Decision fusion methods have also been proposed, such as combining Pauli and Krogager decompositions with local texture descriptors} \cite{papadopoulos2024pixel}, \highlighting{or fusing PolSAR and thermal infrared data using correlated decisions and majority voting} \cite{papadopoulos2024correlated}. \highlighting{Although effective, these methods often rely on handcrafted features and fixed fusion rules, limiting scalability. On the architectural side, lightweight designs using pseudo-3D and depthwise separable convolutions have been introduced to reduce complexity while maintaining performance} \cite{dong2020polsar}. \highlighting{Other approaches leverage polarimetric power features as physically grounded inputs for deep classification networks} \cite{zhang2024deep}. \highlighting{More recently, confidence-aware fusion has been explored to selectively integrate decomposition-based features, addressing instability and noise sensitivity} \cite{han2023trusted}. \highlighting{Together, these trends reflect a shift toward adaptable, hybrid frameworks that combine interpretability and learning efficiency—yet a unified, trainable solution for jointly extracting and fusing real and complex-valued features remains an open need.}

The main contributions of this article are outlined as follows:
\vspace{-0.8em}
\begin{itemize}
    \item This paper introduces a novel lightweight model for PolSAR image classification that combines both real- and complex-valued processing. The model employs 3D convolutional layers in each domain to capture spatial and polarimetric features, effectively utilizing their complementary strengths. Compared to conventional 2D CNNs, this design significantly reduces the number of parameters, lowering the risk of overfitting while improving classification performance.\vspace{-0.8em}
    
    \item \highlighting{Depth-wise convolution is employed to refine spatial features with minimal computational overhead, significantly reducing parameter count and enhancing model compactness.}\vspace{-0.8em}
    \item \highlighting{A coordinate attention mechanism is integrated to emphasize informative regions, improving classification accuracy while maintaining low training and inference costs.}

\end{itemize}

\section{Architecture of the proposed DDF2Pol Model}
\label{sec:model}
\begin{figure*}[!t]
\centering
\includegraphics[width=0.95\linewidth]{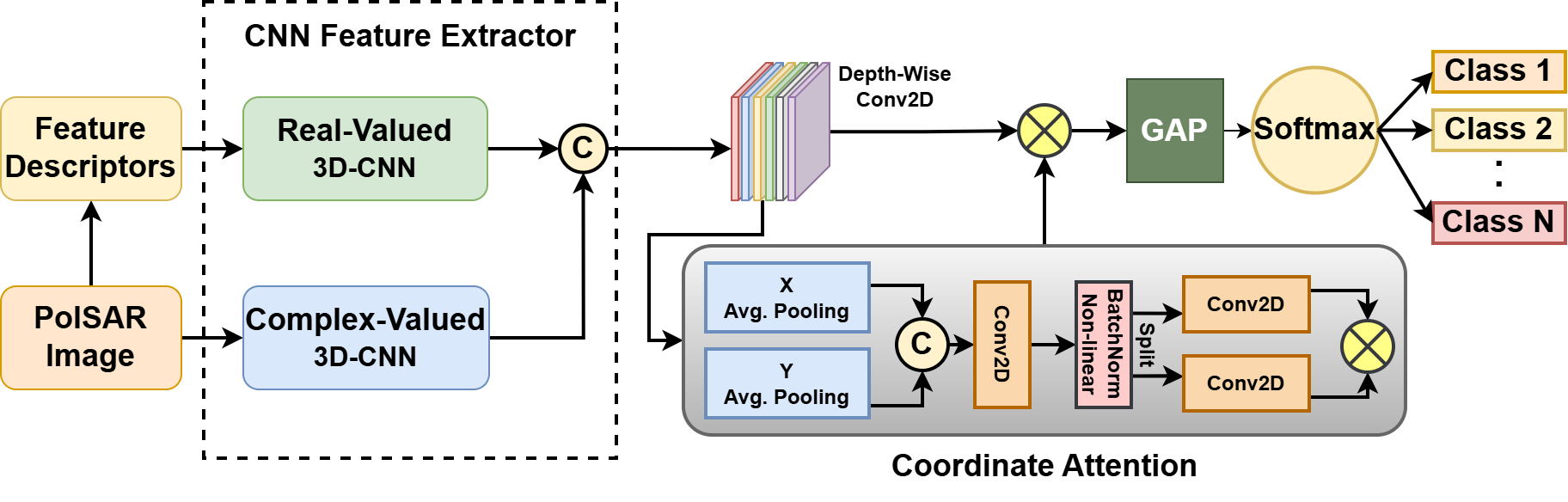}
\vspace{-1em}
\caption{Overall Architecture of the Proposed DDF2Pol Model.}
 \label{fig:model}
\end{figure*}

In this paper, a novel model for PolSAR image classification is introduced, with its detailed architecture illustrated in Fig. \ref{fig:model}. The proposed DDF2Pol framework comprises two parallel streams, a depthwise convolution module for spatial feature enhancement, a coordinate attention block, and a Global Average Pooling (GAP) layer. The first three components serve as feature extractors, whereas classification is performed by the GAP layer.

The process begins by extracting spatial and polarimetric features from both the processed PolSAR data and the feature descriptor images through two parallel streams of 3D convolutional blocks, with each stream consisting of two 3D convolutional layers. The first stream applies real-valued convolutions to process the feature descriptors, while the second stream utilizes complex-valued convolutions to handle the complex PolSAR data. \highlighting{Specifically, six elements from the upper triangular part of the polarimetric coherency matrix are used as complex-valued input to the second stream, while twelve statistical and decomposition-based descriptors derived from the same matrix are used in the first stream.} \highlighting{These descriptors, including SPAN, which represents the total backscattered power, are commonly used in PolSAR classification and help enrich the feature space with physically interpretable information. The complete list of input features is summarized in Table}~\ref{tab:features}.

\begin{table}[t]
\centering
\caption{Polarimetric Descriptor Features Extracted From the Coherency Matrix ($T$)}
\label{tab:features}
\resizebox{\linewidth}{!} {
\begin{tabular}{ll}
\hline
\textbf{Feature} & \textbf{Description} \\  
\hline
$RF1-RF6 = |{T_{ij}}|$ & Magnitude of $T_{ij}, \{i,j\} = 1\rightarrow3$\\
$RF7 = 10 \log_{10}(\text{SPAN})$ & Polarimetric total power \\  
$RF8 = T_{22}/\text{SPAN}$ & Normalized ratio of power $T_{22}$ \\  
$RF9 = T_{33}/\text{SPAN}$ & Normalized ratio of power $T_{33}$ \\  
$RF10 = |T_{12}| / \sqrt{T_{11} \cdot T_{22}}$ & $T_{12}$ relative correlation coefficient \\  
$RF11 = |T_{13}| / \sqrt{T_{11} \cdot T_{33}}$ & $T_{13}$ relative correlation coefficient \\  
$RF12 = |T_{23}| / \sqrt{T_{22} \cdot T_{33}}$ & $T_{23}$ relative correlation coefficient \\  
\hline
\end{tabular}}
\end{table}

\subsection{CNN\highlighting{s} Feature Extractor}
Currently, most PolSAR classification models rely on either real-valued CNN\highlighting{s} (RV-CNNs) \citep{jamali2022polsar, jamali2023local} or complex-valued CNN\highlighting{s} (CV-CNNs) \citep{alkhatib2025polsar, li2024complex} architectures. While RVNNs effectively capture information from feature descriptor images, they struggle to fully utilize the rich polarimetric features present in the original complex-valued PolSAR data. Conversely, CV-CNNs can extract both magnitude and phase information, preserving the full complexity of PolSAR signals. However, this approach is computationally expensive, as it requires nearly twice the number of parameters compared to RVNNs. To address these limitations, a two-branch feature fusion network is proposed, incorporating a real-valued feature extractor using 3D-CNN\highlighting{s} and a complex-valued feature extractor using 3D-CV-CNN\highlighting{s}. Each stream is designed with a compact architecture that utilizes a reduced number of parameters to enhance computational efficiency while preserving essential spatial and polarimetric features. This design improves overall feature extraction performance while maintaining a balance between accuracy and efficiency.

The proposed design utilizes two 3D real-valued convolutional layers with ($3 \times 3 \times 3$) filters for feature extraction. The first layer employs 16 filters, while the second contains 32. The use of smaller kernels enhances computational efficiency while preserving effective feature extraction \citep{pei2022small}. For original PolSAR feature extraction, a similar number of layers and filters are employed, but with 3D complex-valued convolutional filters of the same size ($3 \times 3 \times 3$). These filters enable the simultaneous extraction of spatial and polarimetric features by capturing local spatial dependencies while preserving interactions across the three polarimetric channels.

\begin{figure}[!t]
\centering
\includegraphics[width=0.95\linewidth]{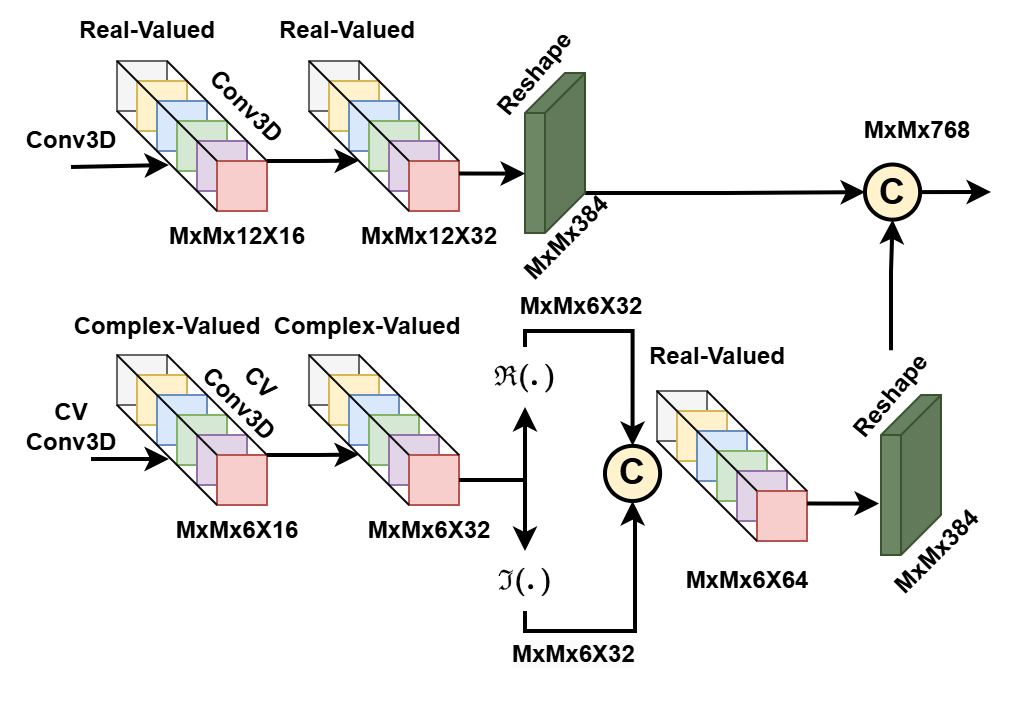}
\vspace{-1em}
\caption{CNN\highlighting{s} Feature Extractor.}
 \label{fig:CNN}
\end{figure}

To integrate complex-valued features into the subsequent processing pipeline, the output from the last layer of the complex-valued stream is transformed into a real-valued representation by concatenating its real and imaginary components. Subsequently, for both streams, the 3D feature cubes are reshaped into 2D feature maps and concatenated, preparing them for spatial feature enhancement in the next layer. Fig. \ref{fig:CNN} shows a detailed diagram of the CNN\highlighting{s} feature extractor

\subsection{Spatial Feature Refinement using Depth-Wise Convolution}
To enhance spatial feature representation while maintaining computational efficiency, a depth-wise 2D convolutional layer is applied following the CNN-based feature extraction stage. Unlike conventional convolution, which aggregates information across channels, depth-wise convolution processes each channel independently \cite{yang2025dsc}, preserving its unique spatial characteristics. This operation reduces computational complexity and parameter count while ensuring that essential spatial features are retained.  

The number of filters in the depth-wise convolution layer is set to match the number of input channels, allowing for consistent spatial refinement across all extracted features. By following this approach, the model effectively enhances spatial representations without introducing significant computational overhead, contributing to improved classification performance.

\subsection{Coordinate Attention (CA)}
\label{sec:CA}

Coordinate Attention (CA) is an efficient attention mechanism that captures both channel-wise and positional information with minimal computational cost. Given an input tensor $\mathbf{X} \in \mathbb{R}^{H \times W \times C}$, the module first encodes spatial information by applying average pooling along the height and width dimensions:

\begin{equation}
x_h = \frac{1}{W} \sum_{i=0}^{W-1} \mathbf{X}(h, i, c), \quad
x_w = \frac{1}{H} \sum_{i=0}^{H-1} \mathbf{X}(i, w, c),
\end{equation}

where $x_h \in \mathbb{R}^{H \times 1 \times C}$ and $x_w \in \mathbb{R}^{1 \times W \times C}$ represent the aggregated features along the width and height, respectively.

Next, $x_h$ and $x_w$ are reshaped and concatenated, forming a unified descriptor along the spatial dimension. A shared $1 \times 1$ convolution $F_s$ followed by a non-linear activation function $\psi(\cdot)$ (typically ReLU) is applied:

\begin{equation}
l = \psi(F_s([x_h, x_w])),
\end{equation}

The result $l$ is then split into two separate tensors, $l^h$ and $l^w$, corresponding to height and width attention branches. These are processed with independent $1 \times 1$ convolutions $F_h$ and $F_w$, followed by a sigmoid activation $\sigma(\cdot)$ to produce the attention weights:

\begin{equation}
g^h = \sigma(F_h(l^h)), \quad
g^w = \sigma(F_w(l^w)),
\end{equation}

The final attention map $\mathbf{W}$ is computed as the outer product of $g^h$ and $g^w$, capturing cross-dimension interactions:

\begin{equation}
\mathbf{W} = g^h \times g^w,
\end{equation}

Finally, the output of the attention module is obtained by applying element-wise multiplication between the input tensor and the attention map:

\begin{equation}
\tilde{\mathbf{X}} = \mathbf{X} \odot \mathbf{W},
\end{equation}

where $\odot$ denotes element-wise multiplication. This operation enhances informative features while suppressing less relevant ones across both spatial dimensions.

\section{Experiments and analysis}
\label{sec:results}

\subsection{Polarimetric SAR Datasets}
Two PolSAR datasets were used to evaluate the performance of the proposed model. The first dataset, the Flevoland image, consists of L-band PolSAR data with four looks \citep{yu2011unsupervised}, featuring a spatial resolution of 12 meters and dimensions of $750 \times 1024$ pixels. Acquired by NASA/JPL’s AIRSAR system on August 16, 1989, over Flevoland, Netherlands, it contains 15 distinct land cover categories \citep{cao2021polsar}. The second dataset, covering the San Francisco region, includes C-band full polarimetric data with four looks \citep{xing2017feature}, collected using the L-band AIRSAR system. It has a spatial resolution of 10 meters and dimensions of $900 \times 1024$ pixels, with classification into five terrain categories \citep{liu2022polsf}. The Pauli RGB maps and their corresponding ground truth maps for these datasets are presented in Figs.~\ref{fig:FL_Results} and \ref{fig:SF_Results}, respectively.

\subsection{Experimental Configuration}
All experiments were conducted using Python 3.9 and TensorFlow 2.10.0 on a Windows 10 machine with 64 GB of RAM and an NVIDIA GeForce RTX 2080 GPU (8 GB VRAM). The Adam optimizer was employed with a learning rate of $1 \times 10^{-3}$ for the both datasets. Training was performed with a batch size of 128 for up to 100 epochs, with early stopping applied if no improvement was observed for 10 consecutive epochs, reverting to the best-performing model weights. To ensure fair comparisons, 1\% of the samples from each dataset were randomly selected for training. The selection process was stratified to maintain an equal number of samples per class for training, mitigating imbalanced representation and ensuring a balanced distribution across all classes.

\subsection{Determining optimal patch size for best performance}
This section analyzes the impact of image patch size on the DDF2Pol model’s performance in PolSAR classification, focusing on identifying the optimal patch size for each dataset. Patch size determines the spatial context extracted, influencing feature representation. Larger patches may incorporate neighboring class information, potentially degrading feature extraction, while smaller patches risk losing essential spatial details. The study evaluates patch sizes \{$5\times5$, $7\times7$, $9\times9$, $11\times11$, $13\times13$, $15\times15$, $17\times17$\} across both datasets. As shown in Table~\ref{tab:opt_window}, the optimal patch size is $15\times15$ for the Flevoland dataset and $13\times13$ for the San Francisco dataset.

\begin{table}[t]
\centering
\caption{The Overall Accuracy (\%) of the proposed DDF2Pol model at various patch sizes.}
\label{tab:opt_window}
\resizebox{0.8\linewidth}{!} {
\begin{tabular}{ccc}
\hline
Patch Size & Flevoland      & San Francisco  \\ \hline
5x5        & 89.79          & 88.06          \\
7x7        & 93.14          & 91.20          \\
9x9        & 95.13          & 92.90          \\
11x11      & 96.62          & 94.54          \\
13x13      & 97.28          & \textbf{95.41} \\
15x15      & \textbf{97.58} & 95.17          \\
17x17      & 97.50          & 95.03          \\ \hline
\end{tabular}
}
\end{table}

\subsection{Ablation Study}
\begin{table}[t]
\centering
\caption{Results of ablation studies on different combinations of model components applied to the Flevoland dataset.}
\label{tab:ablation}
\resizebox{0.9\linewidth}{!} {
\begin{tabular}{lccc}
\hline
Combination & OA (\%) & AA (\%) & Kappa $\times$ 100 \\ \hline
RV          & 94.74   & 94.38   & 93.95              \\
CV          & 95.73   & 95.63   & 95.99              \\
RV+CV       & 96.02   & 96.36   & 95.65              \\
RV+CV+DW    & 96.89   & 97.08   & 96.83              \\
RV+CV+DW+CA & 98.16   & 98.20   & 97.99              \\ \hline
\end{tabular}
}
\end{table}

To evaluate the impact of combining real- and complex-valued domains, as well as the inclusion of depthwise convolution and attention mechanisms, an ablation study \highlighting{is} conducted on the Flevoland dataset. In this study, the Real-Valued \highlighting{network} (RV) consists of two layers of real-valued convolutional blocks, while the Complex-Valued \highlighting{network} follows the same structure but operates in the complex domain. Table~\ref{tab:ablation} presents the results for different model configurations.

The RV-only configuration achieves an OA of 94.74\%, whereas the CV-only counterpart performs better with an OA of 95.73\%, highlighting the advantage of processing PolSAR data using complex-valued CNNs. Integrating both RV and CV streams results in a slight improvement of 0.29\% in OA. Further refinement through depthwise (DW) convolution enhances OA to 96.89\%, demonstrating its effectiveness in improving spatial feature extraction. The addition of the attention mechanism significantly boosts performance, achieving an OA of 98.16\%. This aligns with existing literature \citep{hu2018squeeze, hou2021coordinate}, which establishes that attention mechanisms enhance model performance with minimal computational overhead.

From a computational perspective, the base model (RV+CV) comprises 54,447 parameters, assuming a complete architecture including \highlighting{convolution} blocks and a dense layer for classification. Incorporating depthwise convolution increases the parameter count by 7,680, bringing the total to 62,127. Finally, adding the attention block results in a total of 91,371 parameters. Despite their significant impact on classification accuracy, the combined depthwise convolution and attention mechanisms contribute minimally to the overall model complexity, demonstrating their efficiency in enhancing performance without excessive computational cost.

\subsection{Comparison with Other Methods}
To evaluate the effectiveness of DDF2Pol, it \highlighting{is} compared against six state-of-the-art models: 3D-CNN \citep{zhang2018polarimetric}, WaveletCNN \citep{jamali2022polsar}, PolSARFormer \citep{jamali2023local}, 3D CV-CNN \citep{tan2019complex}, CV-2D-3D \citep{li2024complex}, and HybridCVNet \citep{alkhatib2024polsar}. Among these, 3D-CNN, WaveletCNN, and PolSARFormer are real-valued networks. 3D-CNN was one of the first models to introduce CNNs for PolSAR classification, while WaveletCNN and PolSARFormer are more recent approaches. On the other hand, 3D CV-CNN, CV-2D-3D, and HybridCVNet are complex-valued models. 3D CV-CNN was one of the earliest to apply complex-valued CNNs to PolSAR classification, whereas CV-2D-3D and HybridCVNet represent more recent advancements. These models were chosen to ensure a fair comparison, covering both real-valued and complex-valued approaches, highlighting DDF2Pol’s ability to effectively leverage both data types. \highlighting{Unlike DDF2Pol, HybridCVNet and CV-2D-3D accept only complex-valued inputs. In HybridCVNet, the term "hybrid" refers to the combination of CNNs and ViT blocks for local and global feature extraction, rather than dual-domain fusion. In CV-2D-3D, the attention module is real-valued and processes the real and imaginary components separately. Additionally, both models use standard CV 2D and 3D convolutions, where the filter depth adapts to input channels, potentially leading to over-parameterization. In contrast, DDF2Pol adopts a compact design by introducing depth-wise convolution and employs 3D-CNNs in both real and complex branches for efficient and integrated polarimetric feature extraction.}

\subsubsection{Networks Complexity}

Table \ref{tab:complexity} presents the computational complexity of each model, including the number of parameters, floating point operations (FLOPs), and multiply and accumulate operations (MACs). DDF2Pol stands out as the most efficient, requiring only 91,371 parameters, 7.6M FLOPs, and 2.0M MACs—significantly lower than all other models. In contrast, WaveletCNN and HybridCVNet are the most computationally expensive, with 4.7M and 12.3M parameters and FLOPs exceeding 390M and 29M, respectively. Among real-valued models, PolSARFormer balances complexity with 1.3M parameters and 25.7M FLOPs, while CV-2D-3D represents a higher-cost complex-valued alternative at 2.7M parameters and 5.7M FLOPs. Despite its lightweight design, DDF2Pol integrates depthwise convolution and attention mechanisms, optimizing feature extraction while keeping computational costs minimal. This makes it a highly efficient alternative for PolSAR classification without sacrificing performance.

\begin{table}[t]
\centering
\caption{Parameters, FLOPs, and MACs of each Model used in the research}
\label{tab:complexity}
\begin{tabular}{lccc}
\hline
Model        & Parameters & FLOPs       & MACs        \\ \hline
3D-CNN       & 1,497,515  & 2,895,960   & 1,447,500   \\
WaveletCNN   & 4,714,043  & 391,738,732 & 195,843,446 \\
PolSARFormer & 1,351,961  & 25,750,832  & 12,871,384  \\
3D-CVNN      & 1,871,422  & 1,774,176   & 886,656     \\
CV-2D-3D     & 2,714,889  & 5,675,196   & 2,833,008   \\
HybridCVNet  & 12,312,502 & 29,476,624  & 14,725,512  \\
DDF2Pol & 91,371     & 7,606,272   & 2,045,952  \\
\hline
\end{tabular}
\end{table}

\subsubsection{Analysis of Flevoland Classification Results}

\begin{figure*}[t]
\centering
\includegraphics[width=.85\linewidth]{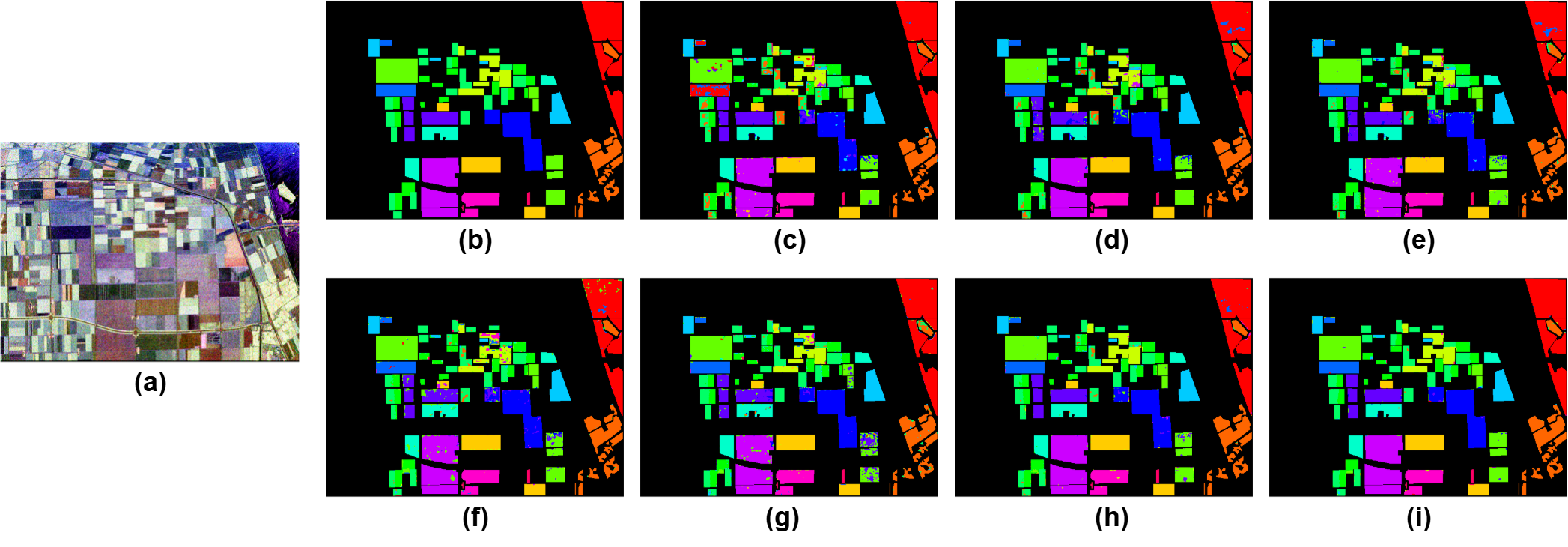}    
 \vspace{-1em}
 \caption{Classification results of the Flevoland dataset. (a) Pauli RGB; (b) Reference Class Map; (c) 3D-CNN; (d) WaveletCNN; (e) PolSARFormer; (f) 3D-CV-CNN; (g) CV-2D-3D; (h) HybridCVNet; (i) Proposed DDF2Pol.}
 \label{fig:FL_Results}
\end{figure*}

Fig.~\ref{fig:FL_Results} presents the classification maps of the Flevoland dataset using different models, while Table~\ref{tab:FL_Results} summarizes their classification performance with only 1\% of the data used for training. The proposed DDF2Pol model achieves the highest accuracy, with an Average Accuracy (AA) of 98.20\%, an Overall Accuracy (OA) of 98.16\%, and a Kappa value of 97.99, outperforming all other methods. Among the real-valued models, PolSARFormer achieves the best results, with an AA of 95.22\% and an OA of 94.41\%, showing an improvement over traditional CNN-based models such as 3D-CNN (AA = 88.47\%) and WaveletCNN (AA = 94.39\%). On the complex-valued side, HybridCVNet performs competitively with an AA of 97.33\%, but DDF2Pol surpasses it by nearly 1\%, demonstrating the effectiveness of combining real-valued and complex-valued features.

A closer look at class-wise accuracy reveals that water and buildings are more easily classified across all models compared to vegetation-based classes. This is likely due to their distinct scattering characteristics, which make them more separable in the feature space. DDF2Pol achieves the highest classification accuracy for the water class at 99.65\% and for buildings at 100.00\%, reinforcing its ability to differentiate these unique classes. Meanwhile, for more challenging vegetation-based classes, such as rapeseed, potatoes, and wheat, DDF2Pol consistently outperforms other models, achieving 97.96\%, 96.70\%, and 98.20\% accuracy, respectively. This suggests that the model effectively captures spatial and polarimetric variations critical for distinguishing between similar vegetation types. The DDF2Pol model  combines real-valued and complex-valued CNNs for feature extraction, followed by depthwise convolution to refine spatial details and an attention mechanism to enhance feature representation. This design allows it to outperform other models by effectively capturing both spatial and polarimetric features. Despite its lightweight structure, DDF2Pol achieves high classification accuracy even with minimal training data, making it a reliable and efficient choice for PolSAR image analysis.

\begin{table*}[t]
\caption{Experimental Results of different methods on Flevoland Dataset.}
\centering

\label{tab:FL_Results}
\resizebox{0.85\linewidth}{!} {
\begin{tabular}{cccc|ccccccc}
\hline
\multirow{2}{*}{Class} & \multirow{2}{*}{Color}                        & \multirow{2}{*}{Train} & \multirow{2}{*}{Test} & \multicolumn{3}{c|}{Real-Valued CNN}                    & \multicolumn{3}{c|}{Complex-Valued CNN}                & \multirow{2}{*}{DDF2Pol} \\ \cline{5-10}
                       &                                               &                        &                       & 3D-CNN & WaveletCNN & \multicolumn{1}{c|}{PolSARFormer} & 3D-CV-CNN & CV-2D-3D & \multicolumn{1}{c|}{HybridCVNet} &                           \\ \hline
Water                  & \cellcolor[HTML]{FF0000} & 138                    & 29,111                & 99.05  & 97.25      & 97.04                             & 95.68  & 98.41      & 97.55                            & \textbf{99.65}                     \\
Forest                 & \cellcolor[HTML]{FF6600} & 138                    & 15,717                & 98.45  & 97.63      & 97.22                             & 99.32  & 93.42      & \textbf{99.60}                             & 99.52                     \\
Lucerne                & \cellcolor[HTML]{FFCC00} & 138                    & 11,062                & 96.97  & \textbf{98.63}      & 97.22                             & 91.38  & 97.49      & 97.81                            & 98.49                     \\
Grass                  & \cellcolor[HTML]{CCFF00} & 138                    & 10,063                & 84.84  & 83.92      & 90.29                             & 72.98  & 87.44      & \textbf{97.84}                           & 95.54                     \\
Rapeseed               & \cellcolor[HTML]{66FF00} & 138                    & 21,717                & 91.23  & 95.65      & 94.48                             & 94.34  & 81.95      & 94.69                            & \textbf{97.96}                     \\
Beet                   & \cellcolor[HTML]{00FF00} & 138                    & 14,569                & 82.55  & 91.54      & 91.76                             & 90.71  & 89.22      & \textbf{93.84}                            & 93.55                     \\
Potatoes               & \cellcolor[HTML]{00FF66} & 138                    & 21,206                & 71.20  & 77.84      & 84.70                             & 87.28  & 94.16      & 90.14                            & \textbf{96.70}                     \\
Peas                   & \cellcolor[HTML]{00FFCC} & 138                    & 10,258                & 98.15  & 96.40      & 98.42                             & 90.40  & 95.56      & 98.81                            & \textbf{99.29 }                    \\
Stem Beans             & \cellcolor[HTML]{00CCFF} & 138                    & 8,333                 & 98.77  & 99.39      & 96.81                             & 98.47  & 98.45      & \textbf{99.60}                             & 99.26                     \\
Bare Soil              & \cellcolor[HTML]{0066FF} & 138                    & 6,179                 & 25.82  & \textbf{98.89}      & 97.97                             & 95.79  & 98.13      & 97.55                            & 98.45                     \\
Wheat                  & \cellcolor[HTML]{0000FF} & 138                    & 17,501                & 85.44  & 96.47      & 90.67                             & 88.57  & 96.42      & 96.45                            & \textbf{98.20 }                    \\
Wheat 2                & \cellcolor[HTML]{6600FF} & 138                    & 10,491                & 98.31  & 84.29      & 95.95                             & 88.50  & 91.81      & \textbf{99.21}                            & 98.55                     \\
Wheat 3                & \cellcolor[HTML]{CC00FF} & 138                    & 21,884                & 97.53  & 99.12      & 97.44                             & 95.12  & 95.10      & 99.59                            & \textbf{99.60}                     \\
Barley                 & \cellcolor[HTML]{FF00CC} & 138                    & 7,231                 & 98.78  & \textbf{99.39}      & 98.29                             & 98.05  & 97.76      & 97.22                            & 98.30                     \\
Buildings              & \cellcolor[HTML]{FF0066} & 138                    & 440                   & \textbf{100.00} & 99.48      & \textbf{100.00 }                           & 98.79  & \textbf{100.00 }    & \textbf{100.00}                              & \textbf{100.00 }                   \\ \hline
\multicolumn{4}{c|}{OA (\%)}                                                                                            & 89.70  & 93.78      & 94.41                             & 92.17  & 93.60      & 96.7676                          & 98.16                     \\
\multicolumn{4}{c|}{AA (\%)}                                                                                            & 88.47  & 94.39      & 95.22                             & 92.36  & 94.35      & 97.3313                          & 98.20                     \\
\multicolumn{4}{c|}{Kappa $\times$ 100}                                                                                        & 88.73  & 93.21      & 93.90                             & 91.45  & 93.01      & 96.4711                          & 97.99                     \\ \hline 
\end{tabular}
}
\end{table*}

\subsubsection{Analysis of San Francisco Classification Results}

\begin{figure*}[t]
\centering
\includegraphics[width=.85\linewidth]{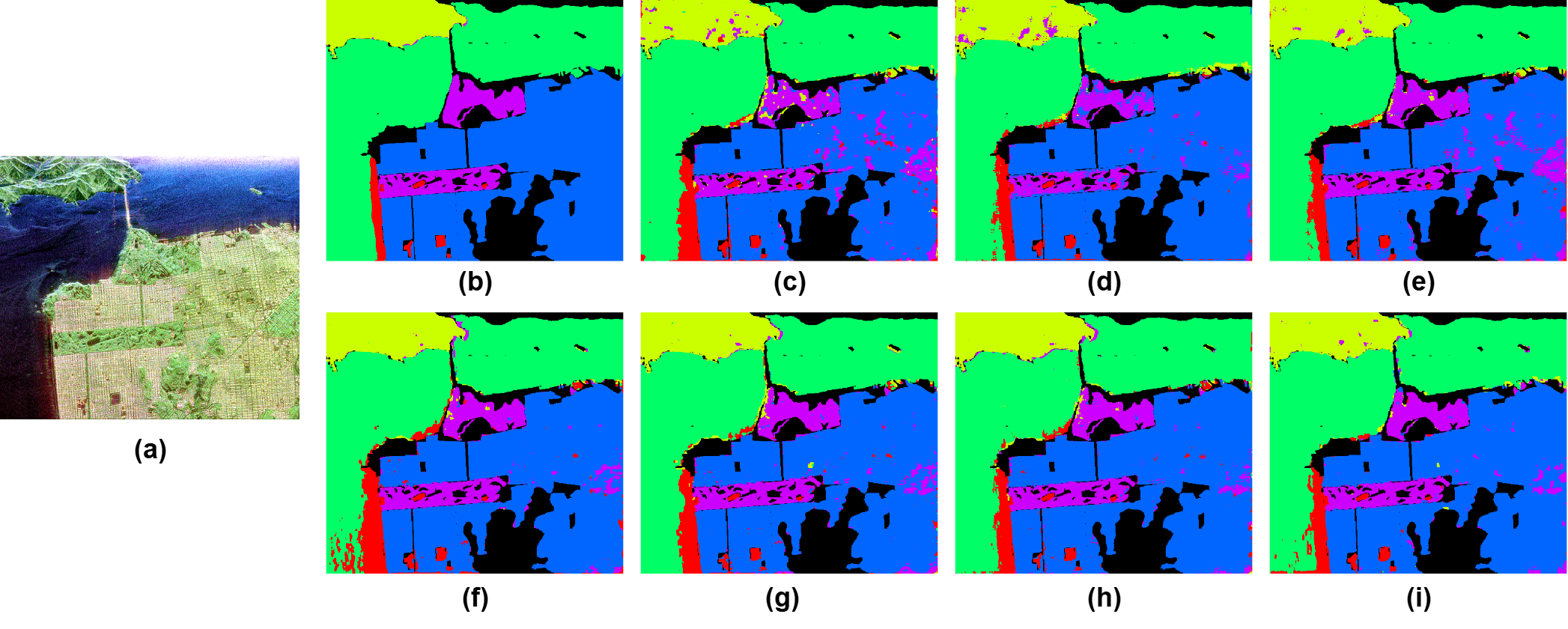}    
 \vspace{-1em}
 \caption{Classification results of the San Francisco dataset. (a) Pauli RGB; (b) Reference Class Map; (c) 3D-CNN; (d) WaveletCNN; (e) PolSARFormer; (f) 3D CV-CNN; (g) CV-2D-3D; (h) HybridCVNet; (i) Proposed DDF2Pol.}
 \label{fig:SF_Results}
\end{figure*}

Fig.~\ref{fig:SF_Results} presents the classification maps generated by different models for the San Francisco dataset, while Table~\ref{tab:SF_Results} summarizes their classification performance using only 1\% of the data for training. The proposed DDF2Pol model achieves the highest overall performance, with an Overall Accuracy (OA) of 96.12\%, an Average Accuracy (AA) of 96.79\%, and a Kappa value of 94.03\%, surpassing all other methods. Among real-valued models, WaveletCNN performs best with an OA of 94.53\%, while among complex-valued models, HybridCVNet reaches 95.96\%. Although DDF2Pol does not achieve the highest accuracy for every individual class, its overall classification performance consistently outperforms other methods, demonstrating its ability to effectively balance accuracy across different land cover types.

A closer look at class-wise accuracy reveals that water and bare soil are more easily classified across all models due to their distinct scattering properties, with DDF2Pol achieving 95.79\% and 99.36\% accuracy for these categories, respectively. While WaveletCNN performs slightly better in urban areas (96.83\% vs. 96.22\%), DDF2Pol remains highly competitive. For vegetation, where classification is more challenging, DDF2Pol maintains a strong performance at 94.47\%, closely aligning with the best-performing models. Comparing Fig.~\ref{fig:SF_Results}(i) with other classification maps, DDF2Pol generates a more refined and consistent classification, reducing misclassifications in complex regions. These results indicate that while some models excel in specific classes, DDF2Pol provides the most balanced and robust classification overall, making it a highly reliable choice for PolSAR image analysis.

\begin{table*}[t]
\caption{Experimental Results of different methods on San Francisco Dataset.}
\centering

\label{tab:SF_Results}
\resizebox{0.85\linewidth}{!} {
\begin{tabular}{cccc|ccccccc}
\hline
\multirow{2}{*}{Class} & \multirow{2}{*}{Color} & \multirow{2}{*}{Train} & \multirow{2}{*}{Test} & \multicolumn{3}{c|}{Real-Valued CNN}                    & \multicolumn{3}{c|}{Complex-Valued CNN}              & \multirow{2}{*}{DDF2Pol} \\ \cline{5-10}
                       &                        &                        &                       & 3D-CNN & WaveletCNN & \multicolumn{1}{c|}{PolSARFormer} & 3D-CV-CNN & CV 2D-3D & \multicolumn{1}{c|}{HybridCVNet} &                           \\ \hline
Bare Soil              &    \cellcolor[HTML]{FF0000}                    & 1,604                  & 12,097                & 98.60  & \textbf{99.39}      & 99.36                             & 99.21  & 97.31      & 97.58                            & 99.36                     \\
Mountain               &       \cellcolor[HTML]{CCFF00}                 & 1,604                  & 61,127                & 93.74  & 92.61      & 97.27                             & 99.21  & \textbf{99.28}      & 99.27                            & 98.09                     \\
Water                  &     \cellcolor[HTML]{00FF66}                     & 1,604                  & 327,962               & 95.12  & 95.34      & \textbf{96.91}                             & 91.70  & 95.04      & 95.15                            & 95.79                     \\
Urban                  &         \cellcolor[HTML]{0066FF}               & 1,604                  & 341,191               & 90.30  & \textbf{96.83}      & 90.44                             & 95.42  & 94.88      & 96.23                            & 96.22                     \\
Vegetation             &     \cellcolor[HTML]{CC00FF}                    & 1,604                  & 51,905                & 82.99  & 75.75      & 90.15                             & 94.82  & \textbf{94.87}     & 94.83                            & 94.47                     \\ \hline
\multicolumn{4}{c|}{OA (\%)}                                                                     & 92.20  & 94.53      & 93.77                             & 94.21  & 95.34      & 95.96                            & \textbf{96.12 }                    \\
\multicolumn{4}{c|}{AA (\%)}                                                                     & 92.15  & 91.98      & 94.83                             & 96.07  & 96.28      & 96.61                            & \textbf{96.79}                     \\
\multicolumn{4}{c|}{Kappa $\times$100}                                                           & 88.16  & 91.51      & 90.51                             & 91.21  & 92.86      & 93.79                            & \textbf{94.03}                     \\ \hline
\end{tabular}
}
\end{table*}

\subsubsection{\highlighting{Models Performance at Different Percentages of Training Data}}

\begin{figure*}[!t]
\centering
\includegraphics[width=.85\linewidth]{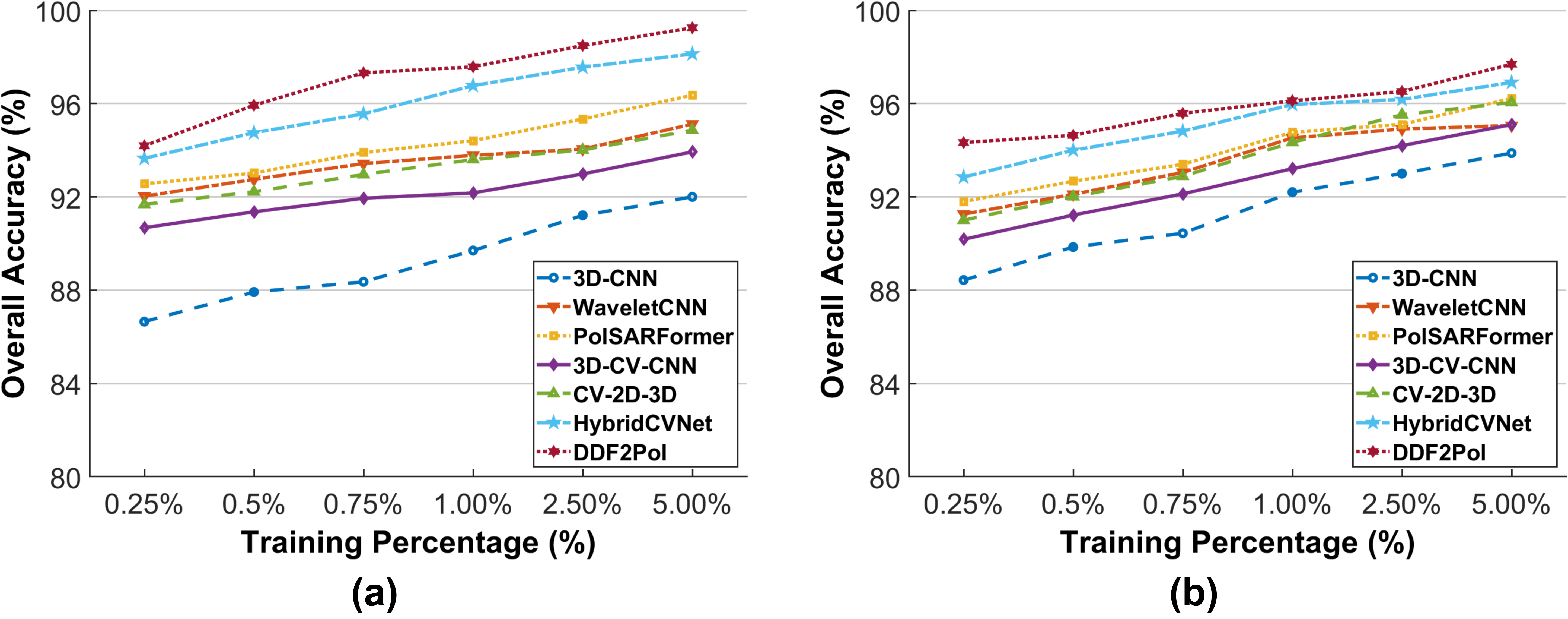}    
 \vspace{-1em}
 \caption{Overall accuracy at different percentages of training data (a) Flevoland (b) San Francisco.}
 \label{fig:Percentages}
\end{figure*}

\highlighting{To assess generalization under limited data, DDF2Pol was compared with competing methods using training ratios from 0.25\% to 5.00\%. As shown in Fig}.~\ref{fig:Percentages},\highlighting{ DDF2Pol consistently outperforms all models across both the Flevoland and San Francisco datasets. With just 0.25\% of labeled data, it achieves 94.19\% and 94.33\% OA on Flevoland and San Francisco, respectively, improving to 99.25\% and 97.69\% at 5.00\%.}

\highlighting{Across all settings, DDF2Pol maintains a consistent lead over real-valued networks (e.g., 3D-CNN, WaveletCNN, PolSARFormer) and complex-valued models (e.g., 3D-CV-CNN, CV-2D-3D, HybridCVNet). The performance gap is especially noticeable under low-data conditions (0.25\%–0.75\%), highlighting the effectiveness of the dual-domain structure and efficient spatial feature refinement. The model also continues to benefit from more training data without signs of overfitting.}

\highlighting{Although only overall accuracy is reported, similar trends were observed in average accuracy and Kappa coefficient. These findings confirm the robustness and scalability of DDF2Pol in real-world scenarios where labeled PolSAR data is scarce or expensive to obtain.}

\section{Conclusion}
\label{sec:conclusion}
This paper introduced DDF2Pol, a lightweight dual-domain CNN model for PolSAR image classification. The architecture integrates real- and complex-valued streams to extract complementary spatial and polarimetric features. Depthwise convolution and coordinate attention are further employed to enhance feature refinement while maintaining computational efficiency. Experiments on two benchmark datasets—Flevoland and San Francisco—demonstrate that DDF2Pol consistently outperforms several state-of-the-art real- and complex-valued models across all tested training ratios. \highlighting{Even when trained with only 0.25\% of labeled samples, DDF2Pol achieves the highest classification accuracy among all compared methods, highlighting its ability to maintain strong performance under severe data scarcity. This result reflects a favorable balance between model complexity and learning capacity in low-data regimes.}

\highlighting{While the proposed framework effectively addresses efficiency-performance trade-offs, several broader challenges in PolSAR classification remain. One key issue is the generalization of models across different sensors, frequency bands, and acquisition conditions. Current models often perform well within specific domains but degrade in cross-domain settings, highlighting the need for robust domain adaptation or meta-learning strategies. Additionally, integrating PolSAR data with complementary modalities such as multispectral imagery, hyperspectral data, or LiDAR presents a promising direction for enhanced scene understanding, albeit with challenges in alignment, resolution mismatch, and multimodal fusion strategies.}

\highlighting{These challenges point toward future research opportunities aimed at developing scalable, adaptable, and generalizable PolSAR classification systems capable of reliable deployment in operational remote sensing applications.}

\section*{Declaration of competing interest}
The author declares that they have no known competing financial interests or personal relationships that could have appeared to
influence the work reported in this paper.

\printcredits

 \bibliographystyle{model1-num-names}

\bibliography{Main_v2}


\end{document}